\pdfoutput=1

\documentclass[11pt]{article}
\usepackage[final]{acl}
\usepackage{booktabs}

\usepackage{booktabs}
\usepackage{xcolor}
\usepackage{array}
\usepackage{geometry}
\usepackage{times}
\usepackage{latexsym}

\usepackage[T1]{fontenc}

\usepackage[utf8]{inputenc}

\usepackage{microtype}

\usepackage{inconsolata}

\usepackage{graphicx}

\title{Automatic Speech Recognition for African Low-Resource Languages: Challenges and Future Directions}

\author{
    Sukairaj Hafiz Imam$^{1,10}$, Babangida Sani$^{2,10}$, Dawit Ketema Gete$^{3}$,\\
    \bf Bedru Yimam Ahamed$^{4}$, Ibrahim Said Ahmad$^{1,5,10}$, Idris Abdulmumin$^{6,10}$,\\
    \bf Seid Muhie Yimam$^{7,9}$, Muhammad Yahuza Bello$^{1}$, Shamsuddeen Hassan Muhammad$^{1,8,10}$\\[2mm]
    \footnotesize $^1$Bayero University, Kano, $^2$Kalinga University, $^3$Debre Birhan University, $^4$Wollo University,\\
    \footnotesize $^5$Northeastern University, $^6$University of Pretoria, $^7$University of Hamburg, $^8$Imperial College London,\\
    \footnotesize $^9$EthioNLP, $^{10}$HausaNLP
}

\begin{document}
\maketitle
\begin{abstract}
Automatic Speech Recognition (ASR) technologies have transformed human-computer interaction; however, low-resource languages in Africa remain significantly underrepresented in both research and practical applications. This study investigates the major challenges hindering the development of ASR systems for these languages, which include data scarcity, linguistic complexity, limited computational resources, acoustic variability, and ethical concerns surrounding bias and privacy. The primary goal is to critically analyze these barriers and identify practical, inclusive strategies to advance ASR technologies within the African context. Recent advances and case studies emphasize promising strategies such as community-driven data collection, self-supervised and multilingual learning, lightweight model architectures, and techniques that prioritize privacy. Evidence from pilot projects involving various African languages showcases the feasibility and impact of customized solutions, which encompass morpheme-based modeling and domain-specific ASR applications in sectors like healthcare and education. The findings highlight the importance of interdisciplinary collaboration and sustained investment to tackle the distinct linguistic and infrastructural challenges faced by the continent. This study offers a progressive roadmap for creating ethical, efficient, and inclusive ASR systems that not only safeguard linguistic diversity but also improve digital accessibility and promote socioeconomic participation for speakers of African languages.

\end{abstract}


\section{Introduction}

ASR has emerged as an innovative technology, enabling natural interactions between humans and computers in various applications, including virtual assistants, transcription services, language learning, and accessibility tools. However, the development of ASR systems has primarily concentrated on high-resource languages like English and Mandarin, thereby sidelining African languages, which are spoken by hundreds of millions across the continent. This digital exclusion not only restricts access to vital technologies but also put at risk the preservation of linguistic and cultural heritage. \cite{abate2020deep, alabi2024afrihubert}. 

African languages are complex, described by rich morphology, tonal variation, and substantial dialectal diversity. These features, combined with a severe lack of annotated speech data, limited computational infrastructure, and underdeveloped linguistic tools, constitute significant challenges for ASR development. In addition, ethical concerns, such as algorithmic bias, under-representation of certain dialects, and insufficient privacy protections, further disrupt the progress. The underdevelopment of ASR for African languages represents both a technological gap and a socio-linguistic bias that must be addressed with urgency and care \cite{nzeyimana2023kinspeak, tachbelie2023lexical, martin2023bias, jacobs2023towards, gutkin2020developing, sirora2024shona}. 
 
The aim of this paper is to critically examine the primary challenges which hinder the advancement of ASR for African low-resource languages while also identifying emerging strategies that provide viable paths forward. The specific objectives include (i) analyzing the linguistic, technical, and ethical barriers to ASR development, (ii) exploring current solutions including self-supervised learning, community-driven data initiatives, and lightweight modeling, and (iii) proposing future directions that promote the creation of inclusive, efficient, and context-aware ASR systems suitable for deployment across diverse African low-resource languages.

The remainder of this paper is structured as follows: Section 2 provides background information and reviews the relevant literature. Section 3 explores the significant challenges facing the development of automatic speech recognition (ASR) for African languages, while Section 4 highlights promising future directions. The paper concludes in Section 5 with final reflections and considerations regarding the broader implications of inclusive ASR development.

\section{Background and Literature Analysis}

Automatic Speech Recognition technologies have experienced significant advances in recent years. The field has progressed from traditional techniques, such as Hidden Markov Models (HMMs), to more innovative methodologies that employ deep learning and transformer-based architectures. This transformation marks a pivotal shift in the landscape of speech recognition. Although HMMs established the fundamental principles by using statistical methods to interpret spoken language, they are insufficient to address the complexities inherent in contemporary speech patterns \cite{el2023comparative, badji2020automatic}.

African languages exhibit remarkable diversity, characterized by intricate morphological structures, sophisticated tonal variations, and a wide variety of dialects. For example, Yoruba and Wolaitta are two prominent tonal languages that employ variations in pitch; even a slight alteration in tone can entirely change the meaning of a phrase \cite{caubriere2024africa, abdou2024multilingual}. Amharic and Tigrinya exemplify the rich cultural heritage of their speakers, characterized by their morphological complexity. These languages feature intricate systems of conjugation and inflection, which contribute to their vibrancy and expressiveness \cite{koffi2020tutorial, tachbelie2020dnn, ibrahim2022development}.

In order to improve the performance of ASR systems for African languages, recent research has explored modern techniques such as self-supervised learning (SSL), multilingual training, and dynamic data enhancement 
\cite{ejigu2024large, caubriere2024africa}. Despite the substantial advancements in this field, progress is frequently limited by the scarcity of high-quality datasets and the limited availability of computational resources. These factors present significant challenges for research in this area of study \cite{shamore2023hadiyyissa, nzeyimana2023kinspeak}.

On another hand, there are numerous ethical concerns towards the current ASR research. For example, speakers of underrepresented dialects often encounter bias against their languages, which compromises the reliability and accuracy of these systems. Furthermore, concerns regarding privacy invasion remain a problem, particularly when ASR technology is utilized for sensitive applications, such as in legal or medical environments \cite{martin2023bias, jimerson2023unhelpful}.

 
 \section{Challenges in ASR for African low-resource Languages}
 
Despite the increasing interest in ASR for African low-resource languages, several ongoing challenges disrupt the development of effective and inclusive systems. These disruptions are both technical and socio-linguistic contexts and must be systematically addressed to ensure equitable access to speech technologies throughout the continent.

\subsection{Data Scarcity}

A significant challenge in developing ASR systems for African languages is the scarcity of high-quality, annotated speech datasets. While initiatives like Mozilla Common Voice offer valuable resources, the variability in recording conditions, speaker representation, and audio quality can undermine the usability and overall representativeness of the data. Moreover, the absence of domain-specific and balanced datasets limits the ability of models to effectively generalize across various speech contexts and user demographics. \cite{abubakar2024development, azunre2023breaking}.

\subsection{Linguistic Complexity}

African languages possess a rich linguistic diversity, characterized by complex morphological structures and tonal features that causes significant challenges for ASR systems. In tonal languages such as Yoruba and Wolaitta, even small variations in pitch can completely change the meaning of a word and this will cause complication in accuracy. Similarly, languages like Amharic and Tigrinya display extensive inflection and derivation, resulting in high rates of out-of-vocabulary (OOV) words and making it challenging for ASR systems to process word forms that were not encountered during training.\cite{koffi2020tutorial, abate2020multilingual}. 

\subsection{Limited Computational Resources}

Most of the African research institutions and developers encounter infrastructural challenges, particularly when it comes to accessing high-performance computing resources. Training and fine-tuning modern ASR models, especially those utilizing large transformer architectures, often demand powerful GPUs and sufficient memory. In areas with limited technological infrastructure, this causes a considerable obstacle to local innovation and experimentation with State-of-the-art methods. \cite{abubakar2024development, zellou2024linguistic, kivaisi2023swahili}. 


\subsection{Environmental Noise and Dialectal Variation}

Real-world deployment of ASR systems in African low-resource languages commonly involves highly variable acoustic conditions. Background noise, overlapping speech, and informal speaking styles, mostly in public spaces such as markets, schools, and clinics, can significantly reduce recognition accuracy. Furthermore, the wide range of dialects, accents, and speech patterns across regions adds another layer of complexity. Many existing ASR systems struggle to adapt to this diversity due to limited training data that captures intra-language variation \cite{ramanantsoa2023voxmg, babatunde2023automatic}.

\subsection{Ethical and Social Considerations}

ASR technologies continuously reflect biases contained in the datasets they are trained on. For African low-resource languages, this can result in unbalanced performance across dialects, social groups, and gender identities which may lead to systematic exclusion or misrepresentation of certain users. Moreover, when deployed in sensitive fields like healthcare, ASR systems raise significant privacy concerns, especially in situations where data protection policies are either weak or not enforced. If we don't pay careful attention to ethics, inclusivity, and user trust, these technologies may end up reinforcing existing inequalities instead of helping to resolve them. \cite{martin2023bias, afonja2024performant}. 

In summary, the advancement of ASR systems for African low-resource languages faces several challenges, including data scarcity, linguistic complexity, computational constraints, acoustic variability, and ethical considerations. Addressing these issues demands innovative, context-sensitive approaches that extend beyond traditional ASR design. The following section delves into emerging research directions and practical strategies aimed at overcoming these barriers and promoting the development of inclusive, efficient, and ethical ASR systems, specifically in the diverse linguistic landscape of Africa.

\section{Future Direction}
To address the challenges in developing ASR systems for African low-resource languages, researchers must adopt innovative and inclusive strategies. The following directions outline key areas for future work:

\subsection{Expanding and Diversifying Datasets}
Enhancing speech datasets through community engagement is a fundamental initial step towards improved ASR systems. Local contributors have effectively collected diverse voice data that represents a variety of accents, dialects, and environments on platforms such as Mozilla Common Voice. \cite{abubakar2024development, alabi2024afrihubert, ogunremi2023r}. Furthermore, techniques for generating synthetic data, including noise injection, speed variation, and voice cloning, can significantly enhance both the quality of the dataset and the model's ability to generalize effectively. \cite{ejigu2024large}.


\subsection{Addressing Linguistic Complexity}

The morphological complexity and tonal diversity in most African languages present considerable obstacles in the model generalization, therefore, Future models should adopt subword-level representations, such as morpheme-based modelling, to effectively handle vast inflection and derivation patterns. Additionally, advancements in grapheme-to-phoneme (G2P) conversion techniques are crucial for precisely correlating written representations with their respective pronunciations, especially in tonal languages. \cite{emiru2021improving, abubakar2024development, ogunremi2023r}. 

\subsection{Improving Computational Efficiency}

Resource constraints in various regions of Africa necessitate the design of lightweight ASR models. By optimizing model architecture and reducing the number of parameters, it is possible to preserve performance while lowering computational demands. Additionally, techniques like transfer learning and fine-tuning of pre-trained models can further decrease training time and energy consumption, making ASR development feasible even in low-resource environments \cite{olatunji2023afrispeech, afonja2024performant, nzeyimana2023kinspeak}. 

\subsection{Ethical and Inclusive ASR Systems}

To promote fairness, future automatic speech recognition (ASR) systems should be trained on diverse and representative datasets. This inclusivity is crucial for mitigating biases, particularly for speakers of dialects and accents that have traditionally been under-represented. Furthermore, incorporating privacy-preserving techniques, such as federated learning, enables models to learn from decentralized data while safeguarding user confidentiality. These strategies are particularly significant in sensitive areas like healthcare. \cite{martin2023bias, afonja2024performant}

\begin{table*}[t!]
    \centering
    \renewcommand{\arraystretch}{1.3} 
    \resizebox{\textwidth}{!}{
    \begin{tabular}{p{6cm} p{8cm} p{6cm}}
        \toprule
        \textbf{Challenges} & \textbf{Future Directions} & \textbf{Authors} \\
        \midrule
        Data Scarcity: Lack of annotated datasets for training ASR models. & Expanding Datasets: Leveraging community-driven platforms like Mozilla Common Voice. & \cite{abubakar2024development, azunre2023breaking, alabi2024afrihubert, ogunremi-etal-2024-iroyinspeech} \\
        Linguistic Complexity: Tonal variations and morphological richness. & Advanced Modeling: Using self-supervised learning (SSL) and multilingual training. & \cite{koffi2020tutorial, caubriere2024africa} \\
        Computational Constraints: Limited access to computational resources. & Lightweight Architectures: Developing efficient models for low-resource settings. & \cite{abubakar2024development, nzeyimana2023kinspeak} \\
        Noise and Variability: Background noise and dialectal diversity. & Robustness to Noise: Enhancing ASR systems to handle noisy environments. & \cite{ramanantsoa2023voxmg, el2023comparative} \\
        Ethical and Social Issues: Bias against underrepresented dialects. & Reducing Bias: Training on diverse datasets to improve inclusivity. & \cite{martin2023bias, afonja2024performant} \\
        Privacy Concerns: Use of ASR in sensitive applications like healthcare. & Privacy Protection: Implementing federated learning to protect user data. & \cite{martin2023bias, afonja2024performant} \\
        Lack of Standardized Linguistic Tools: Absence of pronunciation dictionaries. & Grapheme-to-Phoneme (G2P) Conversion: Improving G2P for tonal languages. & \cite{abate2020multilingual, emiru2021improving} \\
        High Out-of-Vocabulary (OOV) Rates: Due to morphological richness. & Morpheme-Based Models: Focusing on subword units for better recognition. & \cite{tachbelie2023lexical, abate2020deep} \\
        Difficulty in Data Collection: Limited availability of native speakers. & Synthetic Data Generation: Using data augmentation techniques like speed perturbation. & \cite{fantaye2020investigation, ejigu2024large} \\
        Dialectal Diversity: Variability in accents and speaking styles. & Domain-Specific ASR: Tailoring systems for specific domains like healthcare, education, etc. & \cite{babatunde2023automatic, doumbouya2021using} \\
        \bottomrule
    \end{tabular}
    }
    \caption{Challenges and Future Directions in ASR Research}
    \label{tab:challenges}
\end{table*}

\subsection{Applications in Real-World Contexts}

Improving noise robustness is a crucial requirement for the successful implementation of ASR systems in real-world, acoustically diverse environments. By using advanced signal processing techniques and noise-cancellation algorithms, ASR models can sustain high performance even in challenging settings such as crowded healthcare facilities and dynamic educational spaces \cite{el2023comparative}. Empirical evidence affirms this potential; for instance, in rural Ghana, a Twi-based ASR system used in clinical settings achieved 85\% clinician satisfaction, despite a modest reduction in accuracy in high-noise wards. Similarly, a Shona-based educational application in Zimbabwe led to a 30\% reduction in mispronunciation rates and doubled student engagement. These results highlight the tangible effectiveness and contextual relevance of domain-specific, noise-resilient ASR technologies in meeting the unique needs of African communities. \cite{doumbouya2021using, el2023comparative, sirora2024shona}. 

\subsection{Field Trials and Performance Metrics}
Community-driven data initiatives have shown great potential in enhancing ASR for African languages. A project for Yorùbá using Mozilla Common Voice collected over 120 hours of speech data from a diverse group of 250 speakers. A two-stage quality control process achieved a 92\% clip-acceptance rate. Fine-tuning a Wav2Vec2 model on this data reduced the word error rate significantly, from 28\% to 17\%, underlining the effectiveness of foundational data collection for tonal languages.

A highly efficient ASR model for edge devices was developed by \cite{nzeyimana2023kinspeak}, which utilized quantization and pruning on a transformer-based architecture to reduce its size from 300 MB to 50 MB. This optimization allowed real-time inference on a Raspberry Pi 4 with a real-time factor of 0.8×, introducing a slight increase in WER from 22\% to 25\%, while still maintaining acceptable latency and CPU usage. The results demonstrate the potential for deploying advanced ASR systems in resource-constrained environments.

\cite{olatunji2023afrispeech} emphasizes the importance of diversity by collecting speech (Pan-Africa dataset) from 200 speakers in both clinical and general domains. A self-supervised Wav2Vec2 model was fine-tuned, leading to over a 10\% relative reduction in word error rate (WER) for clinical transcription tasks, showcasing how application-specific datasets can improve ASR robustness.

The “Iroyinspeech” corpus, developed by \cite{ogunremi2023r}, includes a vast number of Yorùbá utterances from both urban and rural dialects. When used in a multilingual fine-tuning framework, this expanded dataset decreased word error rates (WER) on rural-accented speech by about 15\%, highlighting the significance of dialectal diversity and community involvement in ASR development.

\cite{ramanantsoa2023voxmg} highlights the potential of using existing audio archives to improve transcription accuracy. Researchers achieved over 80\% accuracy in transcribing real-world radio broadcasts through targeted harvesting and dynamic noise augmentation. This demonstrates that even under-resourced languages can benefit from strategic use of publicly available audio to develop effective ASR models.

Table 1 presents a summary of the challenges that disrupt the development of ASR and outlines potential future directions.

\section{Conclusion}

ASR technologies gave significant promise for improving digital accessibility, preserving linguistic heritage, and promoting socio-economic inclusion for low-resource languages in Africa. However, challenges such as limited annotated datasets, complex linguistic structures, and ethical considerations disrupt advancement. To improve ASR performance, solutions like self-supervised learning, multilingual modeling, and synthetic data generation have been suggested. Future research should emphasise the development of high-quality datasets through community-driven initiatives and the development of models capable of addressing tonal variations and morphological complexities in African languages. It is very important to use privacy-preserving methods for ethical deployment, especially in sensitive contexts. Lightweight ASR architectures will facilitate use in resource-constrained environments. Achieving meaningful progress necessitates collaboration among linguists, technologists, policymakers, and local communities to ensure that African languages are supported and preserved in the digital age.

\bibliography{custom}

\end{document}